\pdfoutput=1

\documentclass[11pt]{article}

\usepackage[preprint]{acl}

\usepackage{times}
\usepackage{latexsym}

\usepackage[normalem]{ulem}
\usepackage{rotating}
\usepackage{tabularray}

\usepackage[T1]{fontenc}

\usepackage[utf8]{inputenc}

\usepackage{microtype}

\usepackage{inconsolata}

\usepackage{graphicx}
\usepackage{subcaption}

\usepackage[ruled,vlined]{algorithm2e}
\usepackage{listings}
\usepackage{multirow}
\usepackage{authblk}

\usepackage{booktabs}

\title{uTeBC-NLP at SemEval-2024 Task 9: Can LLMs be Lateral Thinkers?}

\author[1]{Pouya Sadeghi\thanks{Equal Contribution}}
\author[2]{Amirhossein Abaskohi$^*$}
\author[1,3]{Yadollah Yaghoobzadeh}
\affil[1]{Department of Electrical and Computer Engineering, University of Tehran}
\affil[2]{Department of Computer Science, University of British Columbia}
\affil[3]{Tehran Institute for Advanced Studies, Khatam University, Tehran, Iran}
\affil[1]{\texttt{\{pouya.sadeghi,y.yaghoobzadeh\}@ut.ac.ir}}
\affil[2]{\texttt{aabaskoh@student.ubc.ca}}

\begin{document}
\maketitle
\begin{abstract}
Inspired by human cognition, \citet{jiang-etal-2023-brainteaser} create a benchmark for assessing LLMs' lateral thinking—thinking outside the box. Building upon this benchmark, we investigate how different prompting methods enhance LLMs' performance on this task to reveal their inherent power for outside-the-box thinking ability. Through participating in SemEval-2024, task 9, Sentence Puzzle sub-task, we explore prompt engineering methods: chain of thoughts (CoT) and direct prompting, enhancing with informative descriptions, and employing contextualizing prompts using a retrieval augmented generation (RAG) pipeline. Our experiments involve three LLMs including GPT-3.5, GPT-4, and Zephyr-7B-$\beta$. We generate a dataset of thinking paths between riddles and options using GPT-4, validated by humans for quality. Findings indicate that compressed informative prompts enhance performance. Dynamic in-context learning enhances model performance significantly. Furthermore, fine-tuning Zephyr on our dataset enhances performance across other commonsense datasets, underscoring the value of innovative thinking.\footnote{Our codes and data are publicly available at: \url{https://github.com/Ipouyall/Can-LLMs-be-Lateral-Thinkers}}
\end{abstract}

\section{Introduction}
\label{sec:introduction}

\begin{figure}[htbp]
  \centering
  \includegraphics[width=0.87\linewidth]{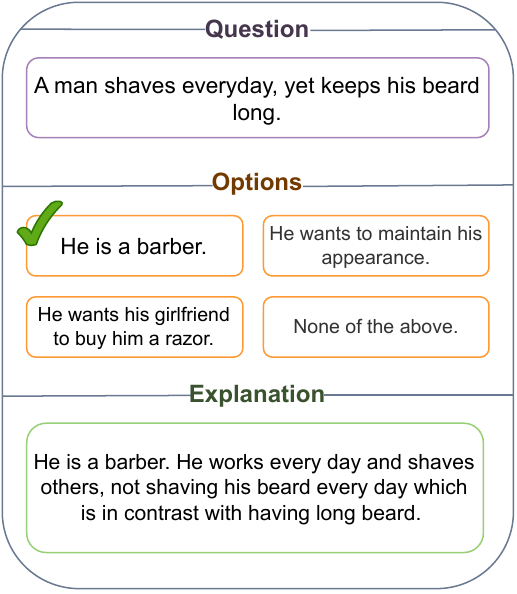}
  \caption{A sample from the sentence puzzle sub-task with an explanation of how this puzzle deprecates default commonsense.}
  \label{fig:sp-sample}
\end{figure}

Human cognition provides the foundational framework for understanding large language model (LLM) development, incorporating two critical thinking modes: vertical and lateral \cite{waks1997lateral}. Vertical thinking, synonymous with logical reasoning, relies on structured analysis and established principles. Conversely, lateral thinking, known for its creativity, challenges conventions and fosters innovative perspectives, enriching language processing capabilities. Recognizing and leveraging the synergy between vertical and lateral thinking are essential in maximizing LLMs' cognitive potential. Integrating both strategies facilitates adaptability and ingenuity in addressing linguistic challenges.

Despite LLMs' success and the abundance of reasoning benchmarks \cite{ho-etal-2023-large, abaskohi-etal-2023-lm, yasunaga2024large}, understanding their reasoning remains incomplete. Many benchmarks prioritize vertical over lateral thinking \cite{waks1997lateral}, inherent in LLMs' pre-training data. \citet{jiang-etal-2023-brainteaser} introduces a challenging dataset, yet thorough analyses of prompting methods are lacking.

Building on previous research examining the impact of prompts on LLMs' performance \cite{webson-pavlick-2022-prompt}, our study aims to validate the genuine lateral understanding capability of LLMs. We participated in SemEval-2024, shared task 9, utilizing various prompts to assess LLMs' lateral thinking abilities in the BrainTeaser multiple-choice QA task \cite{jiang-etal-2023-brainteaser, jiang-ilievski-ma:2024:SemEval2024}. The task focuses on the Sentence Puzzle (see Figure \ref{fig:sp-sample}) and Word Puzzle sub-tasks\footnote{We just participated in the Sentence Puzzle sub-task.}, challenging common sense associations. 

Our team, \textbf{uTeBC-NLP}, employs three different methods to evaluate LLMs' lateral thinking ability: (I) chain of thoughts (CoT)-based strategies, (II) enhancing prompts with a detailed task description and prompt compression, and (III) in-context learning ability, using retrieval-augmented generation (RAG) to select dynamic samples. We conducted these experiments on three LLMs: GPT-3.5\footnote{We used the gpt-3.5-turbo-0125 version.}, GPT-4\footnote{We used the gpt-4-0613 version.}, and Zephyr-7B-$\beta$ \cite{tunstall2023zephyr}, a fine-tuned version of Mistral-7B \cite{jiang2023mistral} trained on a combination of publicly available, synthetic datasets using direct preference optimization (DPO) \cite{rafailov2024direct}.

Our contributions include: (I) exploring the impact of incorporating task information on lateral thinking, (II) developing a thesis-based approach wherein we delineate a path between each question-option pair separately, utilizing this thesis as contextual information in subsequent runs—termed as \textit{external-CoT}, (III) leveraging RAG for generating few-shot examples to assess the efficacy of dynamic few-shot inference (IV) employing the generated thesis context to fine-tune Zephyr-7B-$\beta$ and evaluating its impact on the model's comprehension of commonsense datasets.

In summary, our findings reveal that not all LLMs possess lateral thinking capabilities, with the ability more prominent in models with a greater number of parameters and exposure to extensive data. Proper prompting and introducing unconventional patterns would enhance this capability, by moving beyond conventional linear thinking. Models tend to prefer brief and informative prompts over lengthier alternatives. Notably, we excelled in the Sentence Puzzle sub-task, achieving a remarkable score of 0.975 in solving sentence puzzles, surpassing the baseline of 0.608, and securing the second-highest score.

\section{Background}
\label{sec:background}

\paragraph{Chain of Thoughts Prompting.}
In-context zero-shot and few-shot learning play crucial roles in the success of LLMs. To enhance LLM performance across various tasks, including reasoning tasks, as proposed by \citet{NEURIPS2022_9d560961}, we employ the CoT methodology. Replicating \citet{NEURIPS2022_9d560961}'s setup with sample shots and ensuring their quality can be challenging, so, to ensure fairness and rely on LLMs' knowledge, we adopt \citet{NEURIPS2022_8bb0d291}'s zero-shot-CoT approach and referred to as Simple-Internal-CoT. In our simple-Internal-CoT experiments, we allow LLMs to autonomously handle the problem, thereby separating their performance from the quality of the provided samples. 
We introduced other variations for CoT, (I) Specifiec-Internal-CoT, in which we explain thinking steps to the LLM, and (II) External-CoT, in which we externally help the LLM to follow the steps by asking it to do only one step in each inference, and preserve results use in next steps' prompts.

\paragraph{Enhanced Prompting Strategies.}
Prompt strategies, particularly CoT, prove effective in enhancing LLMs' performance across tasks. However, our experiments demonstrate that CoT may not consistently yield better results. Another approach is boosting the model's performance, using informative prompts, inspired from \citet{fernando2023promptbreeder}. We developed a detailed prompt to familiarize LLMs with diverse riddle approaches in a simple and informative manner, aiming to prevent hasty answers and help them to know how should be faced with riddles. Acknowledging the lengthiness of our detailed prompt and to prevent this factor from limiting LLMs' performance, we created a highly compressed version following \citet{jiang-etal-2023-llmlingua}, retaining essential details to improve LLMs' performance while minimizing prompt length.

\paragraph{In-context Learning.}
In-context learning proved to be one of the most powerful methods to enhance LLMs' performance \cite{NEURIPS2020_1457c0d6}. Few-shot prompting with static samples is examined by \citet{jiang-etal-2023-brainteaser} and shown not to work well. We Employed a RAG pipeline to dynamically select samples from the training split of the dataset. We focused on a three-shot manner and examined different ways of using our RAG pipeline to achieve the best performance. We also explore whether we need to explicitly mention the relation between the question and its answer or whether it can be inferred that effectively.

\section{Methodology}
\label{sec:method}
This section begins with an overview of the dataset, followed by a detailed exposition of our methodology. We will start by elucidating the task information context. Subsequently, we will explore different variations of CoT before concluding with an explanation of our RAG methods.

\begin{figure}[htbp]
  \centering
  \includegraphics[width=\linewidth]{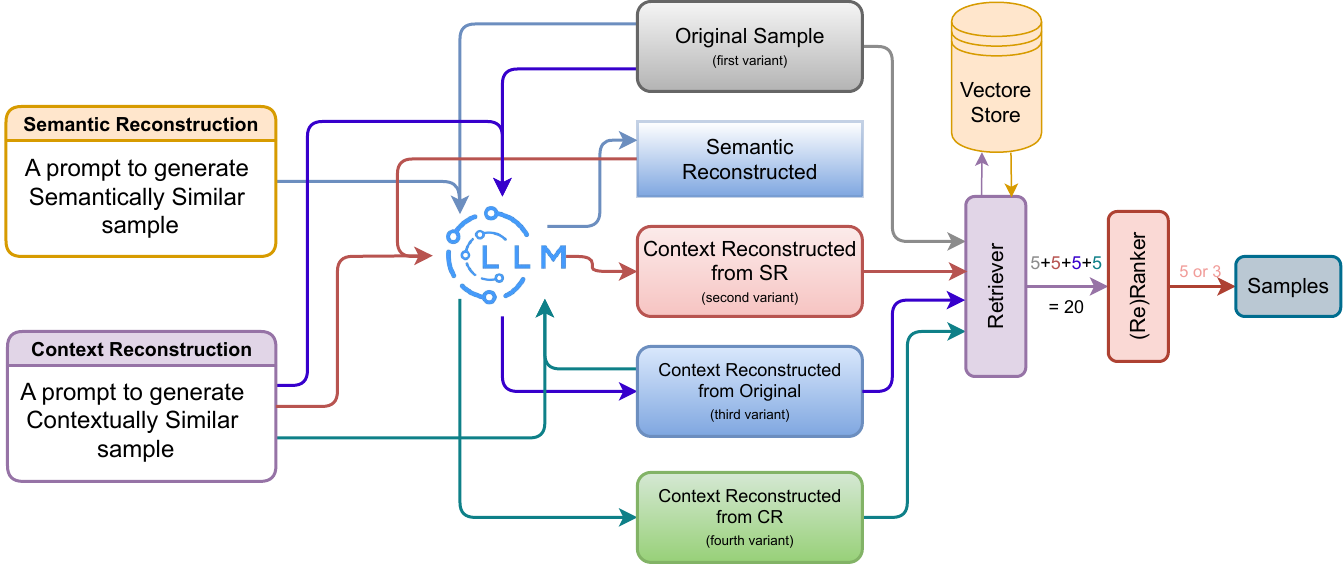}
  \caption{An illustration of our rag-fusion setup. Using an LLM, we generate four variations of the original sample to identify similar ones in the dataset, then rank them to find the closest matches. See appendix \ref{appndx:ragEE} for more details and used prompts.}
  \label{fig:rf-short}
\end{figure}

\begin{figure*}[htbp]
  \centering
  \includegraphics[width=\linewidth]{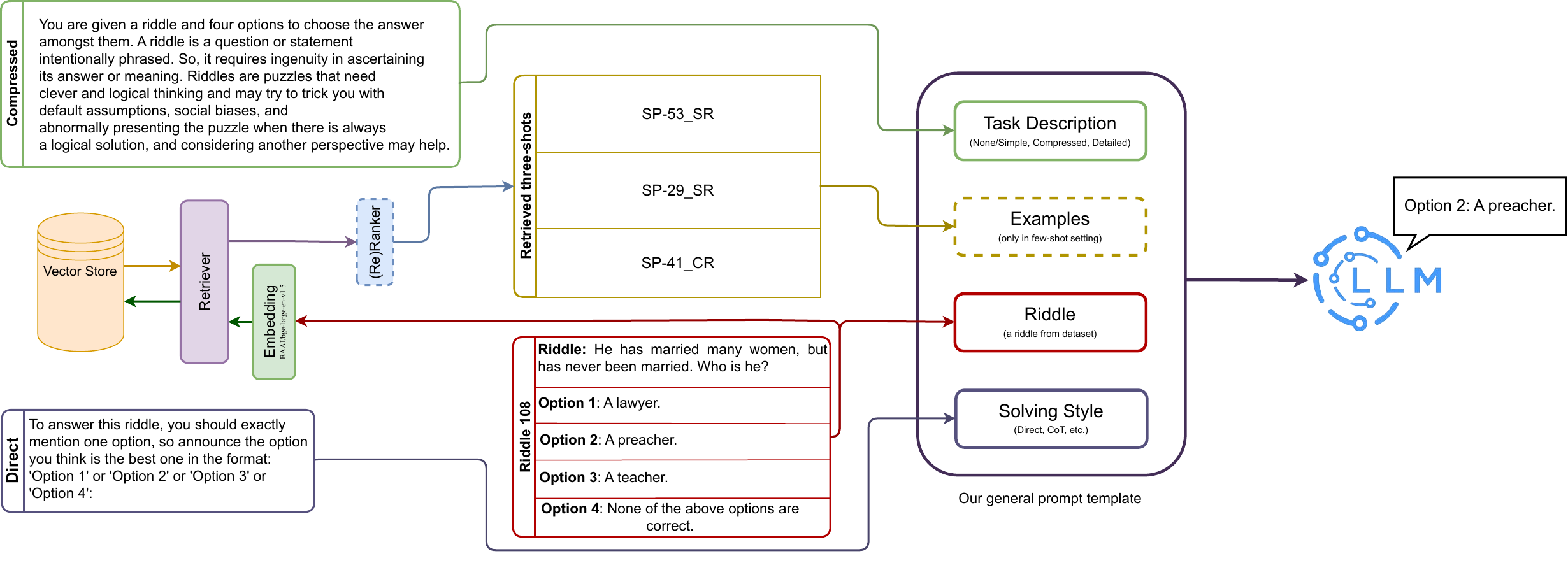}
  \caption{An overview of our approaches in solving the BrainTeaser riddles. In this setup, we have a \textbf{direct} prompt that asks the model to find the appropriate answer. To provide more information to the model, we can offer some task explanation, with the \textbf{compressed} version depicted in this figure. Finally, we utilize our RAG setup to provide the model with in-context examples. In some experiments, we also include the \textbf{theses} for each question-option pair in the prompt, serving as an unbiased link between the question and the option.}
  \label{fig:diagram}
\end{figure*}

\subsection{Dataset}
\label{subsec:dataset}

\paragraph{BrainTeaser.}
The BrainTeaser dataset \cite{jiang-etal-2023-brainteaser} is a multiple-choice question-answering task, designed to evaluate a model's capability for lateral thinking and its ability to challenge default commonsense associations. Created to address the gap in the NLP community's attention towards tasks requiring implicit and intricate reasoning, the dataset relies on human-like commonsense mechanisms. The authors devised a three-step approach to create the first lateral thinking benchmark, involving data collection, distractor generation, and making adversarial examples. They produced 1,100 puzzles with detailed annotations. Assessing models' lateral reasoning consistency, they enhanced BrainTeaser questions with semantic and contextual adjustments. Experiments with top-notch language models showed a significant performance difference from humans, particularly evident across adversarial formats, which aim to avoid cheating in scores by memorizing or previously seen examples. The dataset includes 627 samples for sentence puzzles \footnote{Train: 507, Validation: 120, Test: 120} and 492 samples for word puzzles \footnote{Train: 396, Validation: 96, Test: 96}. In the case of sentence puzzles utilized in our experiments, the average number of tokens in questions is 34.88, with an average of 9.11 tokens in the answers. Our experiments are focused on the Sentence Puzzle sub-task and report our results on test split (post-evaluation phase).

\paragraph{Additional Datasets.}
We generated a dataset based on BrainTeaser's train data that contains a thinking path between a riddle and each of its options, prompting GPT-4 and revised by authors to avoid any bias. Then fine-tuned Zephyr on this dataset. As explored by \citet{jiang-etal-2023-brainteaser}, Fine-tuning models on vertical thinking and traditional commonsense datasets can't improve performance on BrainTeaser and the model's lateral thinking ability. We aim to evaluate whether the fine-tuned model demonstrates an improvement in general commonsense knowledge and examine how lateral thinking ability could affect the model's performance. We utilized SWAG \cite{zellers-etal-2018-swag} and CommonsenseQA \cite{talmor-etal-2019-commonsenseqa} for this purpose, as they don't need lateral thinking.

SWAG is a vast and diverse dataset designed for grounded commonsense inference, comprising over 113,000 sentence-pair completion examples sourced from internet text. Each example presents a context sentence followed by a partial continuation, prompting the selection of the most plausible completion among four choices.

CommonsenseQA (CSQA) serves as a rigorous benchmark for commonsense reasoning, featuring multiple-choice questions requiring an understanding of everyday situations, world facts, and causal relationships. Questions are associated with concepts from a large commonsense knowledge graph, interconnected through various relations, challenging models to engage in complex commonsense reasoning. Our evaluation incorporates 150 random samples from each of these datasets.

\subsection{Task Informative Context}
\label{subsec:task-info}

One approach to evaluating whether LLMs possess lateral thinking abilities is to prompt them explicitly for such capabilities. A key strategy involves providing hints about the task, signaling to the model that it should engage in unconventional thinking. In pursuit of this, we design three variations for task description: \textbf{(I) Simple}, which doesn't provide any special detail and serves as a base to provide evidence of how description could affect the model's performance, \textbf{(II) Detailed}, which would provide detailed information for the task and introduce common tricks to the LLM, and \textbf{(III) Compressed}, which is generated from the detailed variation and it just point out instead of detailed explanation. See Appendix \ref{appndx:prompts} for more details.

\subsection{Thinking Strategy}
\label{subsec:internal-vs-external}

Many different thinking styles are recommended to enhance LLMs' ability to better perform on difficult tasks. CoT prompting has emerged as a potent strategy for enhancing the LLMs' performance, particularly in tasks requiring complex reasoning such as arithmetic and commonsense reasoning \cite{zhang2022automatic, diao2023active, zou2023meta} and known as a popular choice for these complex tasks. However, the question of how CoT should be implemented remains an open challenge. Moreover, we should be aware that CoT won't achieve the best results in all cases, and use it or not, depends on the task and model.

We consider CoT prompting as two main approaches: \textbf{(I) Internal CoT} and \textbf{(II) External CoT}. Internal CoT involves guiding the model through step-by-step thinking or incrementally posing questions to facilitate analytical consideration of each option. Our exploration of internal CoT encompasses two types: (I) Simple, and (II) Specified. In Simple Internal CoT, the model is prompted to think step-by-step without explicit specification of each intermediate step. Specified Internal CoT provides the model with explicitly outlined steps to follow in reaching its answer. Conversely, in External CoT, similar to specified-internal-CoT, we defined steps that the model should pass to reach the final answer, but instead of letting the model control the process, we prompt it to do one step in each inference and use the model's response to generate next prompts till we reach to the final answer. Our suggested intermediate reasoning steps, "\textit{find a path between the question and each answer option and then select the most logical one}," are independent for each question-option pair, and referred to as "\textit{thesis}". Then we would use them as context for each option of the riddle and prompt model to solve the riddle regarding provided contexts. In Section \ref{subsec:exp-prompt-methods} and Figure \ref{fig:thinking-style}, various CoT methods along with direct prompting, are compared.


\subsection{In-context Learning}
\label{subsec:rag}
In this method, we let the model learn the task, using sample(s), known as few-shot prompting. In our few-shot experiments, we individually utilized three samples per question (Figure \ref{fig:diagram}). We observed that employing static samples, as traditionally done in few-shot prompts \cite{NEURIPS2020_1457c0d6}, did not yield a significant performance boost, supporting few-shot results examined by \citet{jiang-etal-2023-brainteaser}. To overcome this limitation, we developed a RAG pipeline to select shots dynamically based on each question. Our experiments involved three RAG methods: Ordinary RAG, Ranked RAG (RAG+ReRanker), and RAG-Fusion. Our experiments are all the same for these approaches and samples are selected from train split. Our shot instances are available with three entities: question, ground-truth answer, and explanation. Our explanations are sampled using GPT4, and they are a logical thinking path from question to answer. We also generated another variant for explanation named Summarized, which compressed long explanations. See Appendix \ref{appndx:ragEE} for more detail.

\paragraph{Ordinary RAG.} Within the ordinary RAG approach, we employed the established RAG framework \cite{NEURIPS2020_6b493230} to generate contextualized representations for the question. The RAG model retrieved relevant passages from a knowledge source, supplying contextual information crucial to the model's decision-making process.

\paragraph{Ranked RAG.} The Ranked RAG approach \cite{NEURIPS2020_6b493230} entailed enhancing the RAG framework by incorporating a reranking mechanism. In this variant, retrieved passages would be reranked based on their relevance to the given question, prioritizing those deemed most relevant. This integration is aimed to enhance the quality and relevance of contextual information provided to the model and GPT-4's contexts had the most positive impact\footnote{Help this method to provide more helpful samples for our purpose.} on this approach.

\paragraph{RAG Fusion.} The RAG-Fusion method \cite{rackauckas2024rag} seamlessly integrates elements from both ordinary RAG and Ranked RAG. In this methodology, we generate three distinct variants derived from the original riddle, which are subsequently input into the RAG pipeline for sample retrieval. After this step, a ranker\footnote{The same reranker used as Ranked RAG.} is employed to prioritize these samples (Figure \ref{fig:rf-short}). This multi-step process is meticulously designed to capture diverse contextual nuances and semantic variations, thereby significantly enhancing the overall effectiveness of the RAG-Fusion method. The most important weakness of this approach is that it is time-consuming as we need many LLM inferences that would take a long time.

\paragraph{Benchmarking.} We designed a benchmark, in which we used samples from the train split, and retrieved 5 unique samples at the end, to observe each variant's performance on different setups. for each retrieved sample from the same group, including the sample itself, the method would give one point for that sample\footnote{Max points for one sample: 3, as we have three variations in each group.}as shown on \ref{tab:rag-benchmart} ordinary RAG can satisfy our needs in a more sample-efficient manner.

\begin{table}[htbp]
    \centering
    \begin{tabular}{l|l}
    \textbf{Metric} & \textbf{Score} (\%)  \\ 
    \hline
    Ori             & 97.5                 \\
    Sem             & 87.5                 \\
    Con             & 82.5                 \\
    Ori\&Sem        & 85.0                 \\
    OriSemCon       & 75.0                 \\
    Overall         & 89.2                
    \end{tabular}
    \caption{Our Final Submission for Sentence Puzzle sub-task. This submission was made by GPT-4 in Simple-Internal-CoT and detailed task description setting.}
    \label{tab:submission-scores}
\end{table}

\section{Experiments and Results}
\label{sec:experiments}
In this section, we report and discuss our results. Table \ref{tab:submission-scores} shows our submission scores during the competition. We first report and discuss our results on different methods (See Appendix \ref{appndx:complete-exp} for full results). Next, we will discuss the effect of fine-tuning on our lateral thinking dataset.

\begin{figure}[htbp]
\centering
    \begin{subfigure}{0.42\textwidth}
        \includegraphics[width=\textwidth]{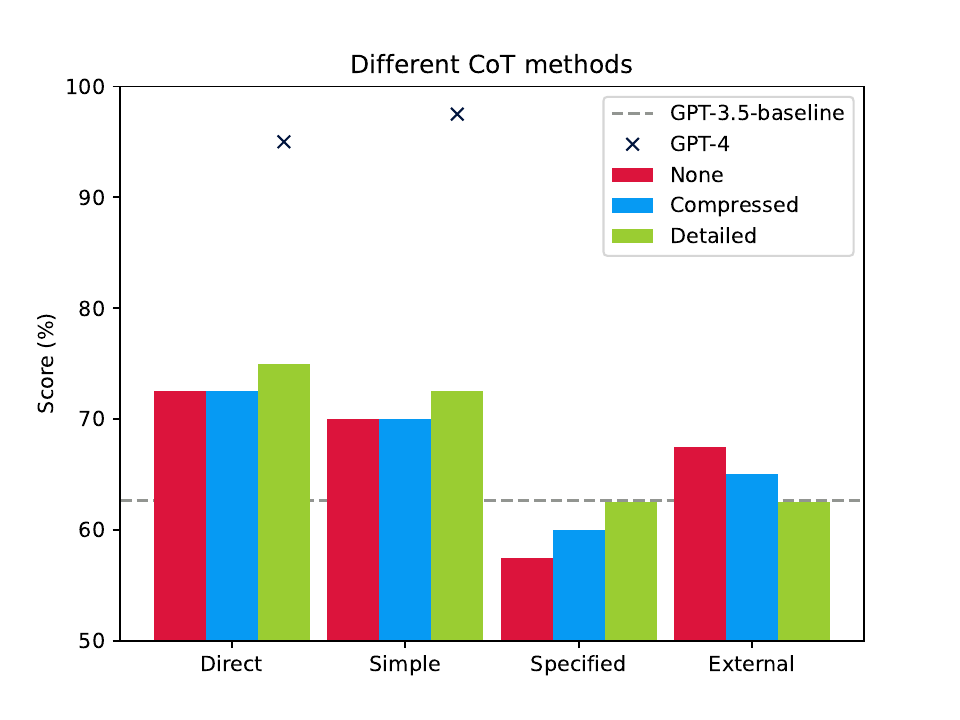}
        \caption{Effect of different thinking (solving) styles on model's performance. Bars belong to GPT-3.5.}
        \label{fig:thinking-style}
    \end{subfigure}
    \hfill
    \begin{subfigure}{0.42\textwidth}
        \includegraphics[width=\textwidth]{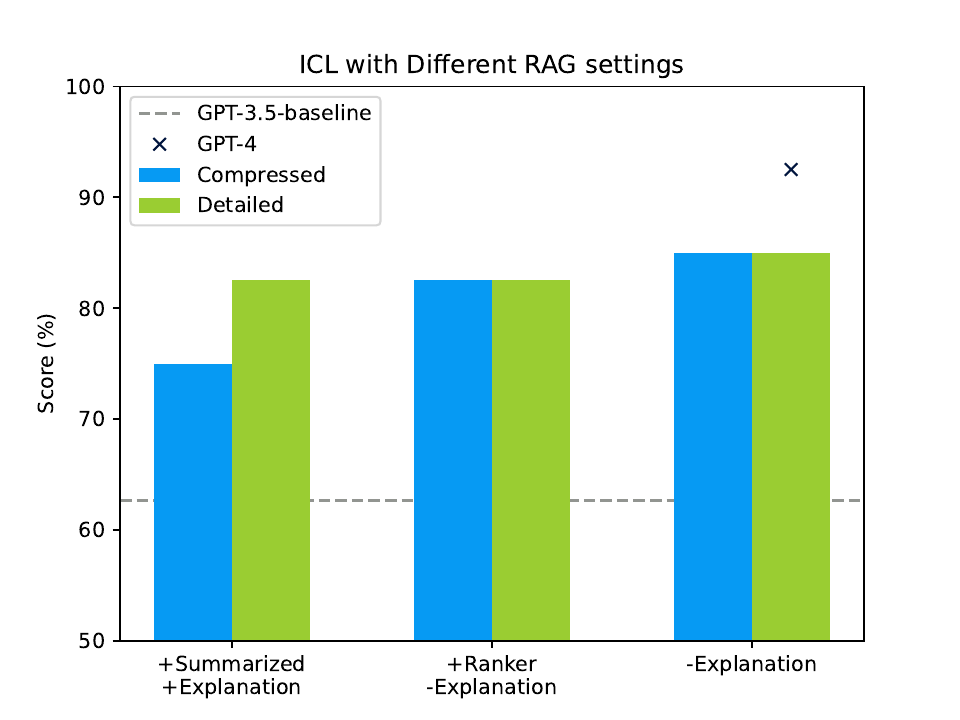}
        \caption{Our proposed RAG-pipelines for dynamic in-context learning(Direct Prompting) and its effect on the model's performance. Bars belong to GPT-3.5.}
        \label{fig:in-context}
    \end{subfigure}        
    \caption{Different prompting approaches and how they affect the model's performance. GPT-3.5-baseline reported by \citet{jiang-etal-2023-brainteaser}.}
    \label{fig:final-result-chart}
\end{figure}

\begin{table}[htbp]
\small
    \centering
    \begin{tblr}{
      cells = {c},
      cell{2}{1} = {r=3}{},
      cell{5}{1} = {r=2}{},
      vline{2-3} = {-}{},
      vline{3} = {3-4,6}{},
      hline{2,5} = {-}{},
    }
    \textbf{Used Samples}                       & \textbf{RAG Type} & \textbf{Hit Rate} \\
    20          & Ordinary          & 0.65              \\
                                                & Fusion            & \textbf{0.767}     \\
                                                & Ranker            & 0.65              \\
    507         & Ordinary          & \textbf{0.753}     \\
                                                & Ranker            & 0.73              
    \end{tblr}
    \caption{Results of RAG's variants on the train split.}
    \label{tab:rag-benchmart}
\end{table}

\subsection{Prompting Methods}
\label{subsec:exp-prompt-methods}
In this section, we detail our exploration of various prompting methods, as outlined in Section \ref{sec:method}. Figure \ref{fig:thinking-style} presents a comparative analysis, revealing that, among different CoT methodologies, simple internal CoT exhibits superior performance. However, it scores lower than direct prompting, without any CoT variation employed. Notably, external CoT outperforms specifying steps in one prompt (specified internal CoT) but falls short compared to simple internal CoT and direct prompting. This is attributed to the impact of prompt length, where longer prompts with similar information weaken performance.

\paragraph{Prompt Length} In task descriptions, providing hints consistently aided the model, with the condensed and detailed versions excelling in different scenarios. Our hypothesis, supported by the results, posits that both prompt and cognitive pathway length significantly influence performance. Extensive factors lead the model to favor concise yet informative prompts, as evidenced by the superior performance with the compressed descriptions.

\paragraph{In-context Learning} As explained in Section \ref{sec:method}, we focused on repeating experiments with Ordinary RAG and RAG+(Re)Ranker in three-shot format, leveraging three entities: question, ground-truth answer, and explanation.

Our explanations, sampled using GPT-4, represent a logical thinking path from question to answer. Additionally, we introduced a summarized variant, generated through Cohere's summarize API \footnote{\url{https://cohere.com/summarize}} \footnote{Settings: length="short", extractiveness="high"} for explanations exceeding 250 words. As depicted in Figure \ref{fig:in-context}, using (Re)Ranker does not significantly enhance performance. Increasing task description details improves performance. Interestingly, we observed that excluding explanation and letting the model infer the thinking path between riddles and their answer will boost LLMs' performance, proving LLMs' ability to extract relations and thinking paths independently. 

Examining three semantically similar questions in a three-shot format, the LLMs' performance converges to a certain score, independent of the task description. The optimal performance is achieved when excluding explanations and using a simple RAG pipeline without using (re)ranker. See Appendix \ref{appndx:ragEE} for more details on settings.

\subsection{Lateral Thinking Tuning}
\label{subsec:exp-lateral-tuning}
Fine-tuning is a core strategy for refining model performance in specific tasks. However, as noted in \citet{jiang-etal-2023-brainteaser}, fine-tuning on other commonsense datasets may not guarantee performance improvements; potentially, it may even lead to a decline in the overall model's performance. This experiment focuses on fine-tuning the model using the dataset generated by GPT-4 and revised by authors, where the model discerns paths between each riddle and its options. The goal is to assess the impact of lateral thinking on model performance by evaluating it on other commonsense datasets. We tried to keep the dataset unbiased, enabling LLM to learn lateral thinking ability.

\begin{table}[htbp]
    \centering
    \begin{tblr}{
      cells = {c},
      cell{2}{1} = {r=4}{},
      cell{2}{2} = {r=2}{},
      cell{4}{2} = {r=2}{},
      vline{2-4} = {-}{},
      vline{4} = {3,5}{},
      vline{3-4} = {4}{},
      hline{2} = {-}{},
      hline{4} = {2-4}{},
    }
    \textbf{Model}                       & \textbf{Dataset} & \textbf{Tuned} & \textbf{Accuracy} \\
    \begin{sideways}Zephyr-7B-$\beta$\end{sideways} & SWAG             & No             & 38                \\
                                         &                  & \textbf{Yes}   & \textbf{46}       \\
                                         & CSQA             & No             & 31.33             \\
                                         &                  & \textbf{Yes}   & \textbf{36}       
    \end{tblr}
    \caption{Fine-tuning experiments, observing model's performance improvement in commonsense.}
    \label{tab:tf-result}
\end{table}

Table \ref{tab:tf-result} showcases that fine-tuning with a lateral thinking approach significantly enhances the model's performance on other commonsense datasets. This result challenges the conventional belief that linear thinking might constrain the model in scenarios requiring unconventional solutions. Adopting an out-of-the-box thinking approach proves beneficial, emphasizing the importance of lateral thinking across diverse contexts.
  
\section{Conclusion}
\label{sec:conclusion}

In conclusion, our study emphasizes the crucial role of prompting methods in augmenting the lateral thinking capabilities of LLMs. Through diverse CoT-based strategies, prompt refinements, and RAG techniques for in-context learning, we showcase the efficacy of well-structured prompts and thinking styles in elevating LLM performance. Additionally, fine-tuning models on a lateral thinking dataset proves advantageous, leading to improved performance on various commonsense tasks. This underscores the significance of integrating out-of-the-box thinking in model training, opening promising avenues for future research to enhance LLMs' reasoning abilities.

\newpage
\bibliography{anthology}

\appendix

\section{Task description Prompts}
\label{appndx:prompts}

\setcounter{table}{0}
\renewcommand{\thetable}{A.\arabic{table}}

During our analysis, we explored various combinations of prompting methodologies. Table \ref{tab:task-desc-prompt} presents our prompts for task description. The initial prompt solely defines what a riddle entails. In the compressed version, we provide general hints, such as avoiding bias. The final prompt includes eleven potential tricks that may occur in the question, along with instructions for the model to evade biases and consider superpower abilities used in the questions. These eleven hints were extracted from other sources.

\begin{table}[!ht]
\centering
\begin{tabular}{p{7.1cm}}
    \multicolumn{1}{c}{\textbf{Prompt}}    \\ \hline
    A riddle is a question or statement intentionally phrased so as to require ingenuity in ascertaining its answer or meaning.                                 \\ \hline
    
    A riddle is a question or statement intentionally phrased so as to require ingenuity in ascertaining its answer or meaning.
Riddles are puzzles that need clever and logical thinking, and may try to trick you with
default assumptions, social biases, and abnormally presenting the puzzle when there are always a logical solution, and considering another perspective may help.
                     \\ \hline
    A riddle is a question or statement intentionally phrased so as to require ingenuity in ascertaining its answer or meaning.
Different ideas can be used in riddles to trick you:
    1. Riddles often employ misdirection, leading you away from the actual solution.
    2. They include elements with double meanings, requiring a keen eye for words with dual interpretations.
    3. Metaphorical wordplay adds another layer, urging you to decipher figurative language.
    4. Look out for exaggeration, as riddles may present overly dramatic details to divert your attention.
    5. Common phrases and sayings may hide within the puzzle, demanding familiarity.
    6. Associations and irony play a crucial role, introducing unexpected connections.
    7. Numerical puzzles can also be part of the mystery, requiring you to decode their significance.
    8. Elemental imagery, drawn from nature, might hold key descriptors.
    9. Rhyming and sound clues can add a poetic dimension.
    10. Avoid sexism and sex cliché, for example, gender bias for jobs, based on their positions or their outcome.
    11. Riddle may try to present something impossible or in contradiction with the reality. Just consider alternative perspectives. \\ \hline
    \end{tabular}
    \caption{Our task description prompts. The first prompt lacks task details, the second is compressed, and the last is detailed, covering all potential tricks.}
    \label{tab:task-desc-prompt}
\end{table}

\section{Generating Path Between Question and Answer}
\label{appndx:path}

One of the methods that was used both in external CoT and step-by-step internal CoT was asking the model to generate a thinking path between a question-option pair. To do that, we asked the model to generate a path between question and option, without giving any judgment on the answer and considering every option can be an answer to the question. To prompt we used to was: "\texttt{Your task is to generate a descriptive explanation from a question to an answer option.
In the following, a question and an option as the answer to the question are provided.
The provided option might or not be a correct answer.
Write a descriptive explanation in at most one paragraph and 200 words to show the thinking path from the question to the option.}." To avoid hallucination, we tried to limit the model's description to 200 by asking the model to do so. Although it does not work all the time, it limits the model's generated words. This limitation would be later beneficial as long input prompts could decrease the model's performance.

\section{Complete experiments and results}
\label{appndx:complete-exp}

\setcounter{table}{0}
\renewcommand{\thetable}{C.\arabic{table}}

\useunder{\uline}{\ul}{}
\begin{table*}[!ht]
\begin{tabular}{l|l|l|l|l}
\textbf{Model}                        & \textbf{Thinking Method}                & \textbf{In-Context Learning} & \textbf{Task Description} & \textbf{Result}     \\ \hline
\multirow{23}{*}{\textbf{GPT 3.5}}    & \multirow{3}{*}{Direct}                 & \multirow{3}{*}{-}           & None                      & 72.5                \\ \cline{4-5} 
                                      &                                         &                              & Compressed                & 72.5                \\ \cline{4-5} 
                                      &                                         &                              & Detailed                  & 75                  \\ \cline{2-5} 
                                      & \multirow{3}{*}{Simple-Internal-CoT}    & \multirow{3}{*}{-}           & None                      & 70                  \\ \cline{4-5} 
                                      &                                         &                              & Compressed                & 70                  \\ \cline{4-5} 
                                      &                                         &                              & Detailed                  & 72.5                \\ \cline{2-5} 
                                      & \multirow{3}{*}{Specified-Internal-CoT} & \multirow{3}{*}{-}           & None                      & 57.5                \\ \cline{4-5} 
                                      &                                         &                              & Compressed                & 60                  \\ \cline{4-5} 
                                      &                                         &                              & Detailed                  & 62.5                \\ \cline{2-5} 
                                      & \multirow{3}{*}{External-CoT}           & \multirow{3}{*}{-}           & None                      & 67.5                \\ \cline{4-5} 
                                      &                                         &                              & Compressed                & 65                  \\ \cline{4-5} 
                                      &                                         &                              & Detailed                  & 62.5                \\ \cline{2-5} 
                                      & Simple-Internal-CoT                     & ES                           & Compressed                & 72.5                \\ \cline{2-5} 
                                      & Direct                                  & ES                           & Compressed                & 75                  \\ \cline{2-5} 
                                      & Direct                                  & ES                           & Detailed                  & 82.5                \\ \cline{2-5} 
                                      & Simple-Internal-CoT                     & ER                           & Compressed                & 72.5                \\ \cline{2-5} 
                                      & Direct                                  & R                            & Compressed                & 82.5                \\ \cline{2-5} 
                                      & Direct                                  & R                            & Detailed                  & 82.5                \\ \cline{2-5} 
                                      & Direct                                  & ord                          & None                      & 85                  \\ \cline{2-5} 
                                      & Direct                                  & ord                          & Compressed                & 85                  \\ \cline{2-5} 
                                      & Direct                                  & ord                          & Detailed                  & 85                  \\ \cline{2-5} 
                                      & Simple-Internal-CoT                     & ord                          & Compressed                & 77.5                \\ \cline{2-5} 
                                      & Specified-Internal-CoT                  & ord                          & Compressed                & 67.5                \\ \hline
\multirow{3}{*}{{\ul \textbf{GPT 4}}} & Direct                                  & -                            & Detailed                  & 95                  \\ \cline{2-5} 
                                      & {\ul \textbf{Simple-Internal-CoT}}      & {\ul \textbf{-}}             & {\ul \textbf{Detailed}}   & {\ul \textbf{97.5}} \\ \cline{2-5} 
                                      & Direct                                  & ord                          & Compressed                & 92.5                \\ \hline
\multirow{9}{*}{\textbf{Zephyr-7B-$\beta$}}      & \multirow{2}{*}{Direct}                 & \multirow{2}{*}{-}           & None                      & 27.5                \\ \cline{4-5} 
                                      &                                         &                              & Detailed                  & 32.5                \\ \cline{2-5} 
                                      & \multirow{7}{*}{Simple-Internal-CoT}    & \multirow{2}{*}{-}           & Compressed                & 37.5                \\ \cline{4-5} 
                                      &                                         &                              & Detailed                  & 15                  \\ \cline{3-5} 
                                      &                                         & ER                           & Compressed                & 40                  \\ \cline{3-5} 
                                      &                                         & ES                           & Compressed                & 42.5                \\ \cline{3-5} 
                                      &                                         & ESR                          & Compressed                & 35                  \\ \cline{3-5} 
                                      &                                         & ord                          & Compressed                & 25                  \\ \cline{3-5} 
                                      &                                         & R                            & Compressed                & 22.5               
\end{tabular}
\caption{Our complete submission result for the post-evaluation phase on test split. In-context learning means using three shots dynamically selected by our RAG's pipeline, in which: \textbf{E}) use Explanation, \textbf{S}) use Summarizer, \textbf{R})use Ranker, and \textbf{ord}) using ordinary rag without explanation and ranker. Our final submission is \uline{underlined}.}
\label{tab:ful-res}
\end{table*}

In this section, we will show our complete results for the sentence puzzle sub-task. Our scores on different experiments are shown in Table \ref{tab:ful-res} and each experiment's descriptions are available in Sections \ref{sec:method} and \ref{sec:experiments}.

\section{RAG extensive experiments}
\label{appndx:ragEE}

\setcounter{table}{0}
\setcounter{figure}{0}
\renewcommand{\thetable}{D.\arabic{table}}
\counterwithin{figure}{section}

\begin{table*}[!ht]
\centering
\resizebox{\textwidth}{!}{
\begin{tabular}{ll}
\multicolumn{1}{c}{}                      & \textbf{Prompt}                                                                                                                                                     \\ \hline
\multicolumn{1}{l|}{\begin{sideways}\textbf{Semantically Related}\end{sideways}} & \begin{tabular}[c]{@{}l@{}}Semantic reconstruction involves rephrasing a given text while preserving its original meaning. \\In this context, you are presented with a riddle. The task is to rephrase the riddle without altering \\the correct answer. Perform a semantic reconstruction of the following riddle.\\ \\ \\ ORG Riddle: "Four men were in a boat on the lake. The boat turns over, \\and all four men sink to the bottom of the lake, yet not a single man got wet! Why?"\\ \\ SR Riddle: "A boat on the lake included four men. All four men on the boat sink to the bottom of \\the lake when it flips over. However, not a single man gets wet! Why?"\\ \\ \\ ORG Riddle: "A plane crashed, and every single person on board this flight was killed, \\yet there were survivors. Explain how?"\\ \\ SR Riddle: "Despite the fact that the entire flight was lost in a plane crash and each single person \\is killed, there were survivors. Describe how?"\\ \\ \\ ORG Riddle: "\{riddle\}"\\ \\ SR Riddle:\end{tabular}                                                                                                                                                                                                                                     \\ \hline
\multicolumn{1}{l|}{\begin{sideways}\textbf{Contextually Related}\end{sideways}} & \begin{tabular}[c]{@{}l@{}}Context reconstruction involves maintaining the original reasoning path while changing \\both the question and the answer to describe a new situational context. In this context, \\you are presented with a riddle. The task is to reconstruct the context of the riddle \\while keeping the original reasoning intact. Ensure that the reconstructed context \\maintains the same reasoning path as the original riddle and also it is reasonable. \\You should change the context and try to avoid it by just replacing some entities and \\trying to convey the question to another scenario. \\Perform a context reconstruction of the following riddle.\\ \\ \\ ORG Riddle: "A woman shoots her husband. Then she holds him underwater for over 5 minutes. \\Finally, she hangs him. But 5 minutes later, they both go out and enjoy a wonderful dinner together. \\How can this be?"\\ \\ CR Riddle: "A woman shoots publicly at people at a National Park. \\The park is full of people, but no one gets killed. How is that possible?"\\ \\ \\ ORG Riddle: "There are 3 apples available for 2 fathers and 2 boys to consume. \\They each receive a single apple. How is it a mathematical possibility?"\\ \\ CR Riddle: "Two mothers and two daughters were asking for new state IDs, \\but the agent only gave out three forms and instructed them on how to fill them out. Why?"\\ \\ \\ ORG Riddle: "\{riddle\}"\\ \\ CR Riddle:\end{tabular} \\ \hline
\end{tabular}
}
\caption{Prompts that are used to generate SR and CR-related samples for the RAG-Fusion method.}
\label{tab:rf-prompt}
\end{table*}

\begin{figure*}[htbp]
  \centering
  \includegraphics[width=\linewidth]{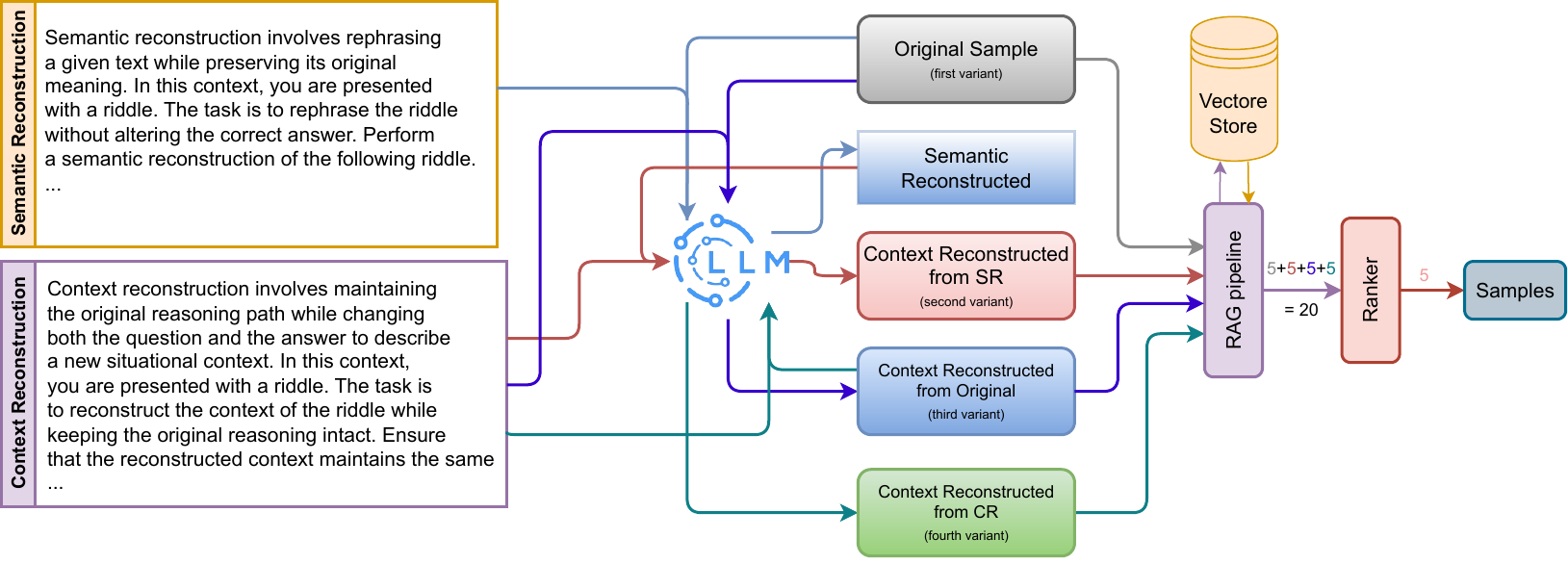}
  \caption{RAG Fusion. The four used variants include: (I) The original riddle, (II) Context reconstruction obtained from semantically reconstructed samples originating from the original riddle, (III) Context reconstruction derived from the original riddle, (IV) Context reconstructed from step 3, then we retrieve similar samples for each variant. In the end, we feed retrieved documents to a ranker to filter them based on similarity and usefulness.}
  \label{fig:rag-fusion}
\end{figure*}

In this section, we discuss our rag experiments in more detail. First, we mention the common setting between different methods, then we will mention each method and explain its specific experimental setup in more detail.

Commonly, we used the Chroma \footnote{\url{https://github.com/chroma-core/chroma}} as our vector store. We employed "bge-large-en-v1.5"\cite{bge_embedding} as our embedding. Our final samples are chosen as the first three unique\footnote{The "unique" term comes meaningful with rag-fusion, as it may retrieve the same samples in each retrieval phase} samples retrieved from our vector store.

Our initial Explanations are selected from the same dataset used for lateral thinking tuning, using ground-truth answers instead of all options. Also for our summarized variant, for explanations longer than 250 words, we used summarizer (in this case, Cohere's summarize API) to summarize explanations.

Overall, our RAG methods can extract different variations in a group, as seen in Table \ref{tab:rag-benchmart} and Origian and SR variations seem to be closely related, but in some cases, it face problems to related CR variation into two other variations.

\paragraph{Ordinary RAG.} We have used it as its ordinary usage. Despite having a close performance to RAG+Ranker, we decided to use this method, reducing the ranker's effect on our performance.

\paragraph{RAG+Ranker} In this method, we first use a normal retriever as Ordinary RAG to retrieve 25 samples from our vector store. Then we fed our query and retrieved 25 samples to reranke(ranker)\footnote{We used Cohere's reranker: \url{https://txt.cohere.com/rerank/}} and kept the first 3 samples with the highest scores. 

\paragraph{RAG Fusion} In this method. We designed two prompts to generate semantically or contextually related (see Table \ref{tab:rf-prompt}). Using those two prompts, by prompting Zephyr-7B-$\beta$, we generate three new variations from the original, and counting the original sample, we feed each variation to the retriever to retrieve 5 samples, which provides 20 samples. Then we would run deduplication to eliminate duplicated samples, which may caused by employing multiple retrieval phases, and rank remained samples using a ranker(re-ranker) and keep the first five samples with the highest score (see Figure \ref{fig:rag-fusion}). The we just continue with the first three samples as our shots.

\end{document}